\documentclass[11pt]{article}
\usepackage[]{acl}
\usepackage{times}
\usepackage{latexsym}
\usepackage[T1]{fontenc}
\usepackage[utf8]{inputenc}
\usepackage{microtype}
\usepackage{graphicx}
\usepackage{booktabs}
\usepackage{placeins}
\usepackage{amsmath,amssymb}
\usepackage{tikz}
\usetikzlibrary{arrows.meta,positioning}
\newcommand{\method}{SERAG}

\title{SERAG: SEmantical and Relational approach for Anomaly detection on knowledge Graphs}

\author{\textbf{Ludovic MONCLA}$^1$, 
\textbf{Antoine SALAZAR}$^1$, 
\textbf{Yassir LAIRGI}$^{1}$, \\
\textbf{Khalid BENABDESLEM}$^1$, 
\textbf{Rémy CAZABET}$^2$  \\
$^1$INSA Lyon, CNRS, UCBL, LIRIS, UMR5205, 69621 Villeurbanne,
France \\
 $^2$Sorbonne Université, CNRS, LIP6, UMR 7606\\
\texttt{\{firstname.lastname\}@liris.cnrs.fr}\\
\texttt{\{firstname.lastname\}@sorbonne-universite.fr}}

\begin{document}
\maketitle
\begin{abstract}
Knowledge graphs constructed from text can contain facts that are individually plausible yet inconsistent with their relational context. We introduce \method{}, a supervised anomaly detector that combines two complementary plausibility estimates: a semantic score computed from separately encoded subject, relation, and object labels, and a relational score obtained through a relational graph convolutional network. The scores are standardized, calibrated, and fused. We evaluate the method on a subset of ICEWS18 with vocabulary-constrained synthetic anomalies generated by an LLM, using pooled and chronological splits. At 10\% anomaly prevalence, \method{} achieves test AUPRCs of 0.321 and 0.258, respectively, compared with 0.159 and 0.146 for the strongest adapted baseline. Fusion improves mean test AUPRC over either branch alone, with a larger gain in the pooled protocol and a smaller validation-selected gain in the chronological protocol. Additional anomaly supervision improves performance. These results support the complementarity of textual and relational signals for detecting synthetic anomalies under the evaluated protocols.

\end{abstract}

\section{Introduction}
Knowledge graphs represent entities and their relations as structured facts, typically triples $(s,p,o)$. They support information retrieval, question answering, and other applications that depend on the quality of the underlying knowledge. When graphs are constructed or enriched from heterogeneous sources, including text, they can acquire erroneous, inconsistent, or atypical facts. Detecting these facts is therefore relevant to the reliability of knowledge-based NLP systems.

Anomaly detection requires more than identifying unusual entity names or rare relations. A fact may be plausible from its textual content but incompatible with the relational regularities of the graph. Conversely, a structurally plausible triple may express a semantically inconsistent relation. Purely structural approaches can overlook information conveyed by entity and relation labels, whereas purely textual approaches do not exploit the relational organization of the knowledge base. This distinction matters especially for candidate anomalies whose individual components remain plausible.

We propose \method{}, a semantic and relational approach for anomaly detection on knowledge graph
\footnote{Code, data splits, cached synthetic anomalies, generation prompts, and experimental configurations are publicly available at \url{https://github.com/ludovicmoncla/serag}.}.
A semantic branch uses frozen Sentence-BERT representations \citep{reimers2019sentence} to estimate compatibility between the subject, relation, and object. 
A relational branch uses a relational graph convolutional network (R-GCN; \citealp{schlichtkrull2018modeling}) to estimate compatibility with the graph neighborhood. We encode the three textual components separately to preserve their roles and fuse the branches after standardization and calibration.

The study asks whether relational context adds useful information to label-based semantic scoring when anomalous facts are constrained to use in-domain entities and relations. ICEWS18 provides a relevant setting: it contains geopolitical events extracted from news, but does not supply anomaly labels. We therefore evaluate on controlled, vocabulary-constrained negatives generated by GPT-4.1. We use both a pooled split and a chronological split, keeping validation and test facts out of the message-passing graph. Although timestamps organize the evaluation, the current model does not explicitly encode temporal coherence.

Our contributions are:
\begin{itemize}
\item A two-branch architecture that makes textual and relational plausibility separately accessible and combines their calibrated scores.
\item A controlled evaluation on shared synthetic anomalies and partitions against adaptations of LoGNet, CCA, and HVSSE, with explicit distinctions between their supervision regimes.
\item Analyses of the contribution of each branch and the effect of available anomaly supervision.
\end{itemize}
The experimental claims concern the supplied ICEWS18 protocol and adapted implementations, rather than state-of-the-art performance across knowledge graph error detection benchmarks.

\section{Related Work}

Knowledge graphs can guide the retrieval and composition of facts for language-based reasoning. Think-on-Graph \citep{sun2024think} organizes iterative graph exploration by a large language model (LLM), while RoG \citep{luo2024rog} separates planning, retrieval, and reasoning. These methods demonstrate the value of explicit graph structure, but their main objective is answering queries rather than detecting anomalous facts.

Other approaches learn representations of triples and their graph context. LoGNet \citep{pirro2022lognet} transforms the knowledge graph into a triple line graph, where each triple becomes a node. A local representation treats each triple as an ordered sequence, while neighboring triples supply global context. This architecture combines information about the components of a fact with information about its relationships to surrounding facts.

Several anomaly detection methods combine multiple views of a knowledge graph. CCA \citep{liu2024knowledge} combines textual and structural views, reconstructs triple entities, and employs adaptive confidence to reduce the influence of noise. HVSSE \citep{xue2024heterogeneous} constructs four views from head and tail co-occurrences and combines local triple encoding with global spatial structure. ProMvSD \citep{zhou2024promvsd} integrates multiple semantic views and prior knowledge obtained with an LLM. Together, these approaches motivate the joint use of semantic and structural information to identify anomalous facts.

Anomaly detection also extends to temporal knowledge graphs, where the evolution of facts provides additional context. C-Unseen \citep{lairgi2026c} studies weak signals in dynamic temporal knowledge graphs through LLM reasoning, considering their coherence and persistence across snapshots.

Within this literature, \method{} focuses on triple plausibility rather than query answering. It builds on approaches combining textual and relational information, with an explicit separation between a local semantic score and an R-GCN-based relational score. Their fusion enables a controlled analysis of the contribution and complementarity of the two branches. The model remains static: chronological evaluation assesses generalization to later observations without explicitly modeling temporal coherence.

\section{Method}

\subsection{Problem Formulation}
A knowledge graph is $\mathcal{G}=(\mathcal{V},\mathcal{R},\mathcal{T})$, with entity set $\mathcal{V}$, relation types $\mathcal{R}$, and observed facts $\mathcal{T}\subseteq\mathcal{V}\times\mathcal{R}\times\mathcal{V}$. Entities are nodes and facts are typed edges. Each entity or relation has a textual representation, such as a label.

For a candidate triple $x=(s,p,o)$, we learn a plausibility score $s_{\mathrm{final}}(x)$ such that coherent facts receive higher scores than anomalies. In the experiments, observed events have label $y=1$ and generated anomalies have label $y=0$. These labels are operational: an observed event is treated as normal, rather than independently verified as true. The model ranks candidate facts or flags low-scoring triples using a validation-selected threshold.

\begin{figure*}[t]
    \centering
    \includegraphics[width=\textwidth]{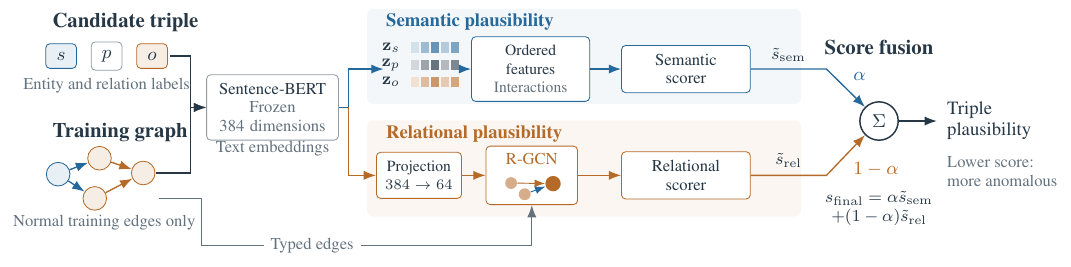}
    \caption{Overview of \method{}. Frozen text embeddings feed two complementary branches: the semantic branch scores ordered triple features and subject--object interactions, while the relational branch uses an R-GCN over normal training edges. Each scoring module includes score standardization and affine calibration before weighted fusion. The separate relation projection and training losses are omitted for clarity.}
    \label{fig:architecture}
\end{figure*}

\method{} combines local compatibility between textual representations and compatibility with relational neighborhoods (Figure~\ref{fig:architecture}). Both branches start from Sentence-BERT embeddings, but only the relational branch performs message passing. Their separation permits an analysis of the information added by graph context beyond textual representations.

\subsection{Semantic Plausibility}
For textual representations $\ell_s$, $\ell_p$, and $\ell_o$, the frozen Sentence-BERT encoder\footnote{We use the pretrained checkpoint \texttt{sentence-transformers/all-MiniLM-L6-v2}.} produces 384-dimensional vectors:
\begin{equation}
\mathbf{z}_q=\operatorname{SBERT}(\ell_q),\quad q\in\{s,p,o\}.
\end{equation}
Separately encoding the components preserves their distinct roles. We form the ordered feature vector
\begin{equation}
\begin{split}
\mathbf{z}_{spo}=[\mathbf{z}_s\Vert\mathbf{z}_p\Vert\mathbf{z}_o
\Vert(\mathbf{z}_s-\mathbf{z}_o)\\
\Vert(\mathbf{z}_s\odot\mathbf{z}_o)],
\end{split}
\end{equation}
where $\Vert$ denotes concatenation and $\odot$ elementwise multiplication. The two interaction terms explicitly expose subject--object differences and similarities. A multilayer perceptron produces
\begin{equation}
s_{\mathrm{sem}}(s,p,o)=\operatorname{MLP}(\mathbf{z}_{spo}).
\end{equation}
It has hidden layers of sizes 512, 128, and 32 with ReLU activations. Dropout rates of 0.4 and 0.3 follow the first two hidden layers.

Ordered concatenation allows asymmetric interactions and distinguishes $(s,p,o)$ from $(o,p,s)$. However, separate encoding does not directly model fine-grained interactions between tokens across the three components. This compact scorer provides a computationally lightweight framework for studying the complementarity of the semantic and relational branches, without assuming that it is the best architecture for semantic scoring.

\subsection{Relational Plausibility}
The relational branch evaluates candidate triples using their typed graph neighborhoods. Entity text embeddings are first projected from 384 to 64 dimensions, yielding the initial representations $\mathbf{h}_i^{(0)}$. A single R-GCN layer then aggregates incoming messages separately for each relation type, while retaining a root transformation and a learned bias. No additional activation function is applied, and dropout with a rate of 0.5 is applied to the resulting representations:

\begin{equation}
\begin{aligned}
\widetilde{\mathbf{h}}_i
={}&
\mathbf{W}_0\mathbf{h}_i^{(0)}
+\mathbf{b}
+\sum_{r\in\mathcal{R}}
\frac{1}{|\mathcal{N}_i^r|}
\sum_{j\in\mathcal{N}_i^r}
\mathbf{W}_r\mathbf{h}_j^{(0)},\\
\mathbf{h}_i
={}&\operatorname{Dropout}
\left(\widetilde{\mathbf{h}}_i; 0.5\right).
\end{aligned}
\end{equation}

Here, $\mathcal{N}_i^r = \{j\mid(j,r,i)\in\mathcal{E}_{\mathrm{train}}^+\}$ denotes the incoming neighbors of entity $i$ through relation $r$, where $\mathcal{E}_{\mathrm{train}}^+$ contains only normal training triples. Edges are directed from subject to object, and no inverse relations are added. Messages are averaged within each relation-specific neighborhood, with empty neighborhoods contributing zero. The matrices $\mathbf{W}_r$ are relation-specific message transformations, $\mathbf{W}_0$ transforms the entity's own representation, and $\mathbf{b}$ is a learned bias. Neither generated anomalies nor validation or test triples are included as message-passing edges.

Relations are projected separately into the same 64-dimensional space. The resulting entity representations $\mathbf{h}_s$ and $\mathbf{h}_o$, together with the relation representation $\mathbf{r}_p$, are used to compute the relational plausibility score:
\begin{equation}
s_{\mathrm{rel}}(s,p,o)=-\frac{1}{d}\|\mathbf{h}_s+\mathbf{r}_p-\mathbf{h}_o\|_1,
\quad d=64.
\end{equation}
This TransE-style scorer \citep{bordes2013translating} maps neighborhood-informed representations to a scalar: a higher value indicates greater compatibility with learned relational regularities. Because the branch is initialized from textual embeddings, it is not a purely topology-only comparator.

\subsection{Standardization, Calibration, and Fusion}
The two branches may produce scores on different scales. For $b\in\{\mathrm{sem},\mathrm{rel}\}$, we standardize the scores using training-set means and standard deviations:
\begin{equation}
u_b(x)=\frac{s_b(x)-\mu_b}{\sigma_b}.
\end{equation}
A learned bias and positive scale then give
\begin{equation}
\tilde{s}_b(x)=\beta_b+\operatorname{softplus}(\gamma_b)u_b(x).
\end{equation}
The positive scale preserves the ordering within each branch. These affine transformations align score scales; they do not establish probabilistic calibration. We fuse the transformed scores as
\begin{equation}
\begin{split}
s_{\mathrm{final}}(x)=\alpha\tilde{s}_{\mathrm{sem}}(x)
+(1-\alpha)\tilde{s}_{\mathrm{rel}}(x),
\end{split}
\end{equation}
with $\alpha\in[0,1]$. The main comparison fixes $\alpha=0.5$; a separate analysis varies its value. The endpoints use only the relational or semantic score in the final decision.

\subsection{Training Objective}
We address class imbalance by averaging binary cross-entropy separately over the two classes and then averaging these class-specific losses. With $\mathcal{D}_c=\{x:y_x=c\}$ and $\sigma_{\mathrm{sig}}$ the sigmoid, this balanced loss is
\begin{equation}
\begin{split}
\mathcal{L}_{\mathrm{bal}}(s,y)=-\frac{1}{2}\Bigg[
\frac{1}{|\mathcal{D}_1|}\sum_{x\in\mathcal{D}_1}\log\sigma_{\mathrm{sig}}(s(x))\\
+\frac{1}{|\mathcal{D}_0|}\sum_{x\in\mathcal{D}_0}\log(1-\sigma_{\mathrm{sig}}(s(x)))\Bigg].
\end{split}
\end{equation}
The fused score and both branches are trained with
\begin{equation}
\begin{split}
\mathcal{L}=\mathcal{L}_{\mathrm{bal}}(s_{\mathrm{final}},y)
+\lambda\big[\mathcal{L}_{\mathrm{bal}}(\tilde{s}_{\mathrm{sem}},y)\\
+\mathcal{L}_{\mathrm{bal}}(\tilde{s}_{\mathrm{rel}},y)\big],
\end{split}
\end{equation}
where $\lambda=0.25$. Auxiliary losses keep each branch discriminative. Differentiable ranking and edge masking are implemented in the source experimental framework but disabled in the reported reference configuration; we make no claim about their effects.

\section{Experimental Setup}

\subsection{Data and Synthetic Anomalies}

ICEWS18 \citep{jin2020recurrent} contains geopolitical events extracted from news and represented as temporal quadruples $(s,p,o,t)$. We use the first 30 daily snapshots, corresponding to timestamps 0 through 696 in increments of 24. This temporal subset contains 44,732 event instances before preprocessing. After deduplicating $(s,p,o)$ triples across snapshots and assigning repeated facts to their earliest occurrence, it comprises 34,064 distinct observed triples involving 8,032 active entities and 216 relation types. These elements are drawn from the complete ICEWS18 vocabulary of 23,033 entities and 256 relations. Both evaluation protocols use this same 30-snapshot subset; timestamps determine the data partitions but are not provided as input to the models. Each experiment therefore trains a single static model whose message-passing graph contains only the normal triples assigned to the training partition.

Since ICEWS18 does not provide anomaly labels, we construct synthetic negative examples with GPT-4.1. For each snapshot, the generator receives up to 100 observed context facts and is instructed to produce events that are deliberately implausible on historical, geographical, or logical grounds. The subject, relation, and object of every generated triple must belong to the vocabulary observed in its source snapshot. Consequently, anomaly generation introduces neither new entities nor new relation types and reduces the possibility of detecting anomalies through trivial out-of-domain cues. We reject duplicate candidates and any generated triple matching an observed positive fact in any of the 30 selected snapshots. Accepted anomalies are cached and reused unchanged across methods, random seeds, and experimental conditions.

These constraints make the generated examples in-domain but do not guarantee that every candidate is factually false or equally difficult for all models. We therefore describe them as synthetic anomalies rather than expert-verified errors. Observed ICEWS18 events are operationally treated as normal, and the reported experiments measure discrimination between these events and the accepted outputs of this particular generation procedure.

\subsection{Adapted Baselines}
Our LoGNet reimplementation retains BiGRU sequence encoding and global enrichment on a triple line graph. For scalability, it uses a sparse line graph and replaces the original semantic attributes with an attention mechanism.

CCA and HVSSE are independent adaptations. For CCA, Sentence-BERT embeddings of entity labels replace descriptions unavailable in ICEWS18; multi-view reconstruction, contrastive alignment, and adaptive confidence are retained. HVSSE retains four co-occurrence views, a BiLSTM encoder, a graph Transformer, and contrastive and translational objectives.

All methods use the same example manifests, seeds, and evaluation metrics. \method{} and LoGNet use training anomaly labels, whereas CCA and HVSSE do not use these labels in their training losses. Labeled validation data nevertheless selects models and thresholds for all methods. The comparison therefore controls examples and partitions, but does not equalize supervision or reproduce the original implementations and benchmark settings.

\subsection{Evaluation Protocols}
\paragraph{Pooled.}
The first 30 snapshots are aggregated and triples are randomly split into 60\% training, 20\% validation, and 20\% test data with class stratification. This evaluates interpolation within an aggregated distribution.

\paragraph{Chronological.}
Snapshots 0--17 provide training data, snapshots 18--23 provide validation data, and snapshots 24--29 provide test data, without temporal shuffling. This yields contiguous 60\%/20\%/20\% temporal partitions and evaluates generalization to later observations. Both protocols therefore use the same 30 snapshots and differ in whether their triples are randomly or chronologically assigned to partitions.

Anomaly prevalence is $N_{\mathrm{anom}}/(N_{\mathrm{normal}}+N_{\mathrm{anom}})$. The main comparison uses 10\% anomalies in each partition. The supervision analysis varies training prevalence over 1\%, 2.5\%, 5\%, 10\%, and 20\%, holding validation and test prevalence at 10\%.

\subsection{Optimization and Metrics}

Each configuration is run on CPU with five random seeds (41--45). \method{}, LoGNet, and HVSSE are trained for up to 300 epochs with an early-stopping patience of 25 epochs. \method{} and LoGNet use Adam with learning rates of $5\times10^{-4}$ and $10^{-2}$, respectively. HVSSE uses AdamW with a learning rate of $10^{-3}$ and 128-dimensional representations. CCA uses AdamW with a learning rate of $10^{-4}$, batches of 256 triples, a maximum of 20 epochs, and a patience of 8 epochs.
For all methods, checkpoint selection and early stopping are based on validation AUPRC. None of the runs reaches its epoch limit. CCA stops after 9--10 epochs, with its best validation AUPRC attained within the first two epochs. \method{} exhibits an initial increase in validation AUPRC followed by a plateau, whereas HVSSE shows more variable learning trajectories across seeds.

The decision threshold maximizes anomaly F1 on validation data. We report test anomaly precision, recall, and F1 at this threshold, plus AUROC and AUPRC with anomalies as the evaluation target. Since larger model scores indicate normality, anomaly ranking uses the reverse ordering. AUPRC is computed by trapezoidal integration of the precision--recall curve. At 10\% prevalence, random-ranking AUPRC is 0.10. Reported values are means and standard deviations across five seeds; we do not treat these descriptive statistics as a significance test.

\section{Results and Analysis}
\subsection{Comparison with Adapted Baselines}

Table~\ref{tab:main} reports the main comparison. \method{} has the highest mean AUROC, AUPRC, precision, and F1 in both protocols. LoGNet has the highest mean recall in the pooled protocol, while HVSSE has the highest mean recall in the chronological protocol.

\begin{table*}[t]
\centering
\small
\setlength{\tabcolsep}{3pt}
\begin{tabular}{llccccc}
\toprule
Protocol & Method & AUROC & AUPRC & Precision & Recall & F1\\
\midrule
Pooled & LoGNet (reimpl.)
 & $0.659\pm0.015$ & $0.159\pm0.010$
 & $0.152\pm0.010$ & $\mathbf{0.555\pm0.096}$
 & $0.237\pm0.013$ \\
Pooled & CCA (adapted)
 & $0.617\pm0.006$ & $0.145\pm0.003$
 & $0.135\pm0.013$ & $0.537\pm0.182$
 & $0.210\pm0.004$ \\
Pooled & HVSSE (adapted)
 & $0.568\pm0.017$ & $0.125\pm0.010$
 & $0.121\pm0.013$ & $0.551\pm0.171$
 & $0.194\pm0.010$ \\
Pooled & \method{}
 & $\mathbf{0.786\pm0.003}$ & $\mathbf{0.321\pm0.011}$
 & $\mathbf{0.308\pm0.016}$ & $0.451\pm0.055$
 & $\mathbf{0.364\pm0.012}$ \\
\midrule
Chronological & LoGNet (reimpl.)
 & $0.632\pm0.018$ & $0.146\pm0.007$
 & $0.138\pm0.006$ & $0.597\pm0.147$
 & $0.223\pm0.011$ \\
Chronological & CCA (adapted)
 & $0.597\pm0.005$ & $0.142\pm0.005$
 & $0.126\pm0.007$ & $0.542\pm0.120$
 & $0.203\pm0.006$ \\
Chronological & HVSSE (adapted)
 & $0.525\pm0.014$ & $0.108\pm0.004$
 & $0.104\pm0.002$ & $\mathbf{0.699\pm0.154}$
 & $0.180\pm0.006$ \\
Chronological & \method{}
 & $\mathbf{0.737\pm0.003}$ & $\mathbf{0.258\pm0.008}$
 & $\mathbf{0.262\pm0.028}$ & $0.389\pm0.069$
 & $\mathbf{0.308\pm0.008}$ \\
\bottomrule
\end{tabular}
\caption{Test results with 10\% synthetic anomalies in every partition (mean $\pm$ standard deviation across five seeds). Baseline labels identify adapted implementations. Bold values mark the best mean in each protocol.}\label{tab:main}
\par\medskip

\begin{tabular}{llccccc}
\toprule
Protocol & Method & 1\,\% & 2.5\,\% & 5\,\% & 10\,\% & 20\,\% \\
\midrule
Pooled & LoGNet
 & $0.184\pm0.014$ & $0.208\pm0.009$
 & $0.168\pm0.008$ & $0.159\pm0.010$
 & $0.170\pm0.008$ \\
Pooled & \method{}
 & $\mathbf{0.231\pm0.005}$ & $\mathbf{0.257\pm0.004}$
 & $\mathbf{0.292\pm0.005}$ & $\mathbf{0.321\pm0.011}$
 & $\mathbf{0.382\pm0.015}$ \\
\midrule
Chronological & LoGNet
 & $0.164\pm0.007$ & $0.155\pm0.006$
 & $0.150\pm0.008$ & $0.146\pm0.007$
 & $0.156\pm0.005$ \\
Chronological & \method{}
 & $\mathbf{0.215\pm0.006}$ & $\mathbf{0.225\pm0.006}$
 & $\mathbf{0.237\pm0.007}$ & $\mathbf{0.258\pm0.008}$
 & $\mathbf{0.289\pm0.016}$ \\
\bottomrule
\end{tabular}

\caption{Test AUPRC as a function of training anomaly prevalence. Validation and test prevalence remain 10\%. Values are means $\pm$ standard deviations across five seeds. LoGNet denotes the adapted reimplementation.}
\label{tab:ratio}
\end{table*}

In the pooled setting, \method{} reaches an AUROC of $0.786$ and an AUPRC of $0.321$, compared with $0.659$ and $0.159$ for LoGNet, the strongest baseline on these metrics. Under chronological evaluation, \method{} reaches $0.737$ AUROC and $0.258$ AUPRC, exceeding LoGNet by 0.105 and 0.112 in absolute terms. Its mean AUPRC is more than three times the random-ranking level in the pooled setting and more than 2.5 times that level in the chronological setting.

At the validation-selected thresholds, \method{} achieves the highest mean precision and F1 in both protocols, with F1 scores of $0.364$ in the pooled setting and $0.308$ in the chronological setting. LoGNet obtains the highest mean recall in the pooled protocol ($0.555$), while HVSSE does so in the chronological protocol ($0.699$). However, their lower precision results in lower F1 scores than \method{}. LoGNet provides the strongest baseline F1 under chronological evaluation ($0.223$). Overall, \method{} provides the best precision--recall balance as measured by anomaly F1.

Chronological evaluation lowers mean AUPRC for all four methods: from 0.321 to 0.258 for \method{}, from 0.159 to 0.146 for LoGNet, from 0.145 to 0.142 for CCA, and from 0.125 to 0.108 for HVSSE. Because both protocols now use the same 30 snapshots, these differences are not confounded by temporal coverage or total source data; they reflect the change from random interpolation to temporally ordered partitions, including the resulting distribution shift and partition composition. Comparisons across methods should also account for their different supervision regimes: \method{} and LoGNet use training anomaly labels, whereas CCA and HVSSE do not use these labels in their training losses.

\subsection{Amount of Anomaly Supervision}
We vary the proportion of labeled training anomalies for the two supervised methods, keeping validation and test prevalence at 10\%. CCA and HVSSE are excluded because their losses do not use anomaly labels; varying their training mixture would address sensitivity to contamination rather than the benefit of negative supervision.

Table~\ref{tab:ratio} shows increasing \method{} performance with more labeled anomalies. From 1\% to 20\%, pooled AUPRC increases from 0.231 to 0.382 and chronological AUPRC from 0.215 to 0.289. At 20\%, its F1 reaches $0.405$ and $0.338$, respectively. Even at 1\% training prevalence, its mean AUPRC exceeds LoGNet's best mean across the tested ratios in both protocols.

LoGNet behaves non-monotonically: its best AUPRC occurs at 2.5\% training prevalence in the pooled setting and at 1\% in the chronological setting. In the latter, performance decreases through 10\% before partially recovering at 20\%. More negative supervision therefore benefits \method{} in this experiment but does not guarantee improvement across supervised architectures. Since changing prevalence also changes the training mixture, these results should be read as sensitivity to the tested supervision setting rather than a general sample-efficiency law.

\subsection{Complementarity of the Branches}

\begin{table*}[t]
\centering\small
\setlength{\tabcolsep}{4pt}

\begin{tabular}{lccccc}
\toprule
Protocol & $0$ & $0.25$ & $0.50$ & $0.75$ & $1.00$ \\
\midrule
Pooled
 & $0.257\pm0.007$ & $0.290\pm0.005$
 & $\mathbf{0.321\pm0.011}$ & $0.307\pm0.005$
 & $0.275\pm0.004$ \\
Chronological
 & $0.173\pm0.005$ & $0.210\pm0.007$
 & $0.258\pm0.008$ & $\mathbf{0.276\pm0.003}$
 & $0.252\pm0.007$ \\
\bottomrule
\end{tabular}

\caption{Test AUPRC for different fusion weights $\alpha$, with 10\% anomalies (mean $\pm$ standard deviation across five seeds). The main comparison uses $\alpha=0.5$.}
\label{tab:alpha}
\end{table*}

We vary $\alpha$ at 10\% training anomaly prevalence (Table~\ref{tab:alpha}). For each random seed, a separate model is trained for every value of $\alpha$.
At $\alpha=0$, only the relational score contributes to the final decision; at $\alpha=1$, only the semantic score contributes. The relational endpoint still uses text-derived entity representations, and the reference objective contains auxiliary losses. Thus, these endpoints assess the contribution of each scoring branch rather than a text-free structural system.

In the pooled protocol, the relational and semantic endpoints reach mean AUPRCs of 0.257 and 0.275. Fusion achieves 0.321 at $\alpha=0.5$, an absolute gain of 0.046 over the semantic endpoint. In the chronological protocol, the endpoints reach 0.173 and 0.252, while the highest mean test AUPRC is 0.276 at $\alpha=0.75$, an absolute gain of 0.024 over the semantic endpoint. Validation AUPRC selects $\alpha=0.5$ in both protocols; for the chronological split, this validation-selected setting obtains a test AUPRC of 0.258, only 0.006 above the semantic endpoint. The higher test value at $\alpha=0.75$ is therefore reported as a sensitivity result rather than a validation-selected improvement. The best F1 is obtained with $\alpha=0.5$ in the pooled protocol and $\alpha=0.75$ in the chronological protocol.

Overall, the semantic branch provides the stronger individual signal in both protocols. Combining it with relational context yields a clear improvement in mean test AUPRC in the pooled setting, from 0.275 to 0.321. In the chronological setting, the validation-selected fusion yields a smaller increase, from 0.252 to 0.258. These results suggest that the benefit of fusion depends on the evaluation setting: complementarity is more pronounced under pooled evaluation, while the modest chronological gain requires further investigation to establish its robustness.

\section{Conclusion}

We presented \method{}, a knowledge graph anomaly detector that combines separately encoded textual components with relation-aware graph propagation. 
On vocabulary-constrained, LLM-generated anomalies in ICEWS18, fusion provides a clear improvement in mean AUPRC over either branch alone under pooled evaluation. The validation-selected gain under chronological evaluation is smaller and requires further investigation to establish its robustness.
At 10\% anomaly prevalence, \method{} achieves AUPRCs of $0.321$ and $0.258$ in the pooled and chronological protocols, respectively. LoGNet is the strongest baseline in both settings, with AUPRCs of $0.159$ and $0.146$, yielding absolute gains of 0.162 and 0.112 for \method{}. Additional anomaly supervision improves its performance in the tested range. Future work should evaluate naturally occurring errors with human annotations, diversify anomaly generators, model temporal coherence explicitly, and investigate richer cross-component semantic interactions.

\section*{Limitations}
The evaluation uses accepted outputs from one LLM generator and temporal subsets of one dataset. Vocabulary constraints and positive-fact filtering do not ensure that every generated event is false or representative of naturally occurring errors. No expert annotation study is reported. Observed events are treated as normal, although extraction errors or unusual but correct events may occur. Performance can reflect generator-specific patterns as well as semantic or relational inconsistency. The results should not be generalized to all geopolitical anomalies or other knowledge graph domains.

\method{} and LoGNet use anomaly labels during training, while CCA and HVSSE do not use them in their losses. The use of labeled validation data also means that the latter methods are not evaluated in a fully label-free pipeline. Architectural adaptations, optimizer choices, and training budgets differ. Consequently, the comparisons establish performance for the reported implementations and protocol, rather than an exact reproduction or a definitive ordering of the original methods.

The message-passing graph excludes anomalous training edges. This assumes access to a clean normal training subgraph, which may not be available in a deployment containing unknown errors. Positive training facts are themselves present as propagation edges, whereas validation and test facts are not. The reference configuration disables edge masking. Further evaluation is needed to quantify sensitivity to this train--test asymmetry and to graph contamination. 
Although the pooled and chronological protocols use the same 30 snapshots and exclude cross-partition triple duplication, the effects of assigning repeated facts to their earliest occurrence, along with unseen entities and snapshot-level distribution changes, require further characterization. The model itself is static and does not represent temporal consistency.

Frozen text embeddings and separate component encoding may miss fine-grained semantic interactions, while the translational scorer imposes a restricted geometric model. Results across five seeds characterize initialization variability but do not establish statistical significance or robustness across datasets.

\section*{Ethical Considerations}
The dataset represents geopolitical events and actors. Implausibility judgments made by a generator or detector can reflect linguistic, geographic, or political biases. Low model plausibility is not evidence that an event is false, and anomaly scores should not be used as autonomous judgments about people, organizations, or countries. Human verification is necessary before consequential action.

Generated negative examples are deliberately unreliable and should be identified as synthetic when released, together with the generation procedure and filtering rules. Data redistribution and access to hosted generation services must follow the applicable dataset and service terms. We make no claim that these constraints have been independently audited in this study.

\bibliography{references}
\end{document}